%% file: AMAI2026_39.tex
\documentclass[runningheads]{llncs}
\usepackage[T1]{fontenc}
\usepackage{graphicx,verbatim}
\usepackage{multirow}
\usepackage{amsmath}
\usepackage{tcolorbox}
\usepackage{booktabs}
\usepackage{adjustbox}
\usepackage{hyperref}
\usepackage{pgfplots}
\usepackage{makecell}
\tcbuselibrary{breakable,skins}
\pgfplotsset{compat=1.18}

\newtcolorbox{reasoningbox}[1][]{
  enhanced,
  colback=gray!5, 
  colframe=gray!60!black, 
  coltitle=white,
  fonttitle=\bfseries\sffamily,
  fontupper=\fontsize{8}{10}\selectfont,
  boxed title style={colback=gray!60!black, rounded corners},
  drop fuzzy shadow, 
  arc=3mm, 
  boxrule=0.5pt,
  left=15pt, right=15pt, top=15pt, bottom=10pt,
  #1
}

\begin{document}
\title{Learning Diagnostic Reasoning for Decision Support in Toxicology}
\titlerunning{Learning Diagnostic Reasoning in Toxicology}
%
\author{Nico Oberländer\inst{1}\thanks{Equal Contribution} \and
David Bani-Harouni\inst{1,2}\protect\footnotemark[1] \and
Tobias Zellner\inst{3} \and
Nassir Navab\inst{1,2} \and
Florian Eyer\inst{3} \and
Matthias Keicher\inst{1,2}}
\authorrunning{N. Oberländer, D. Bani-Harouni et al.}
%
\institute{Computer Aided Medical Procedures, Technical University of Munich, Germamy \and
Munich Center for Machine Learning (MCML), Germany \and
Department of Clinical Toxicology and Poison Control Center Munich, TUM Klinikum rechts der Isar, Germany}
  
\maketitle              
\begin{abstract}
Acute poly-substance intoxication requires rapid, life-saving decisions under substantial uncertainty, as clinicians must rely on incomplete ingestion details and nonspecific symptoms. Effective diagnostic reasoning in this chaotic environment requires fusing unstructured, non-medical narratives (e.g. paramedic scene descriptions and unreliable patient self-reports or known histories), with structured medical data like vital signs. While Large Language Models (LLMs) show potential for processing such heterogeneous inputs, they struggle in this setting, often underperforming simple baselines that rely solely on patient histories. To address this, we present DeToxR (Decision-support for Toxicology with Reasoning), the first adaptation of Reinforcement Learning (RL) to emergency toxicology. We design a robust data-fusion engine for multi-label prediction across 14 substance classes based on an LLM finetuned with Group Relative Policy Optimization (GRPO). We optimize the model's reasoning directly using a clinical performance reward. By formulating a multi-label agreement metric as the reward signal, the model is explicitly penalized for missing co-ingested substances and hallucinating absent poisons. Our model significantly outperforms its unadapted base LLM counterpart and supervised baselines. Furthermore, in a preliminary clinical validation study, the model indicates a clinical advantage by achieving higher micro-F1 (0.644 vs 0.473) and recall in identifying the correct poisons. These results demonstrate the potential of RL-aligned LLMs to synthesize unstructured pre-clinical narratives and structured medical data for decision support in high-stakes environments. Our code is available at \url{https://github.com/nicoberlaender/DeToxR}.


\end{abstract}

\section{Introduction}
\input{sections/01_introduction}

\section{Methodology}

\input{sections/02_methodology}

\section{Experiments}
\input{sections/03_experiments}

\section{Results and Discussion}

\input{sections/04_results}

\section{Conclusion}
\input{sections/05_conclusion}

\begin{credits}
\subsubsection{\ackname}
The authors gratefully acknowledge the financial support by the
Bavarian Ministry of Economic Affairs, Regional Development and Energy (StMWi)
under project ThoraXAI (DIK-2302-0002).

\subsubsection{\discintname}
The authors have no competing interests to declare that are relevant to the content of this article. 
\end{credits}

%
%
%
\bibliographystyle{splncs04}
\bibliography{AMAI2026_39}

\end{document}

%% file: sections/01_introduction.tex
Acute drug intoxication is a common problem in emergency care, requiring rapid identification of likely used substances and timely initiation of a targeted treatment to prevent serious complications~\cite{kulling1986role}. Clinicians must base their diagnostic and therapeutic decisions on a mix of structured data (e.g. symptoms, vital signs) and unstructured non-medical narratives (e.g. paramedic scene descriptions or patient histories and self-reports). In many cases, the diagnosis has to be made under substantial uncertainty: exposure histories may be incomplete or inaccurate, patients may be unable to provide reliable and detailed information, and symptoms can be nonspecific~\cite{yong2025artificial} or masked due to co-ingestion of multiple toxic agents~\cite{otten2012goldfrank}. Laboratory analysis can provide definitive answers, however, the results are usually not obtainable within the therapeutic window, meaning clinicians have to act based on the available information. These characteristics make emergency toxicology a uniquely challenging setting for computational decision support.\looseness=-1

\begin{figure}[t]
    \centering
    \includegraphics[width=\linewidth,clip,trim=1.5cm 2.5cm 2cm 2cm]{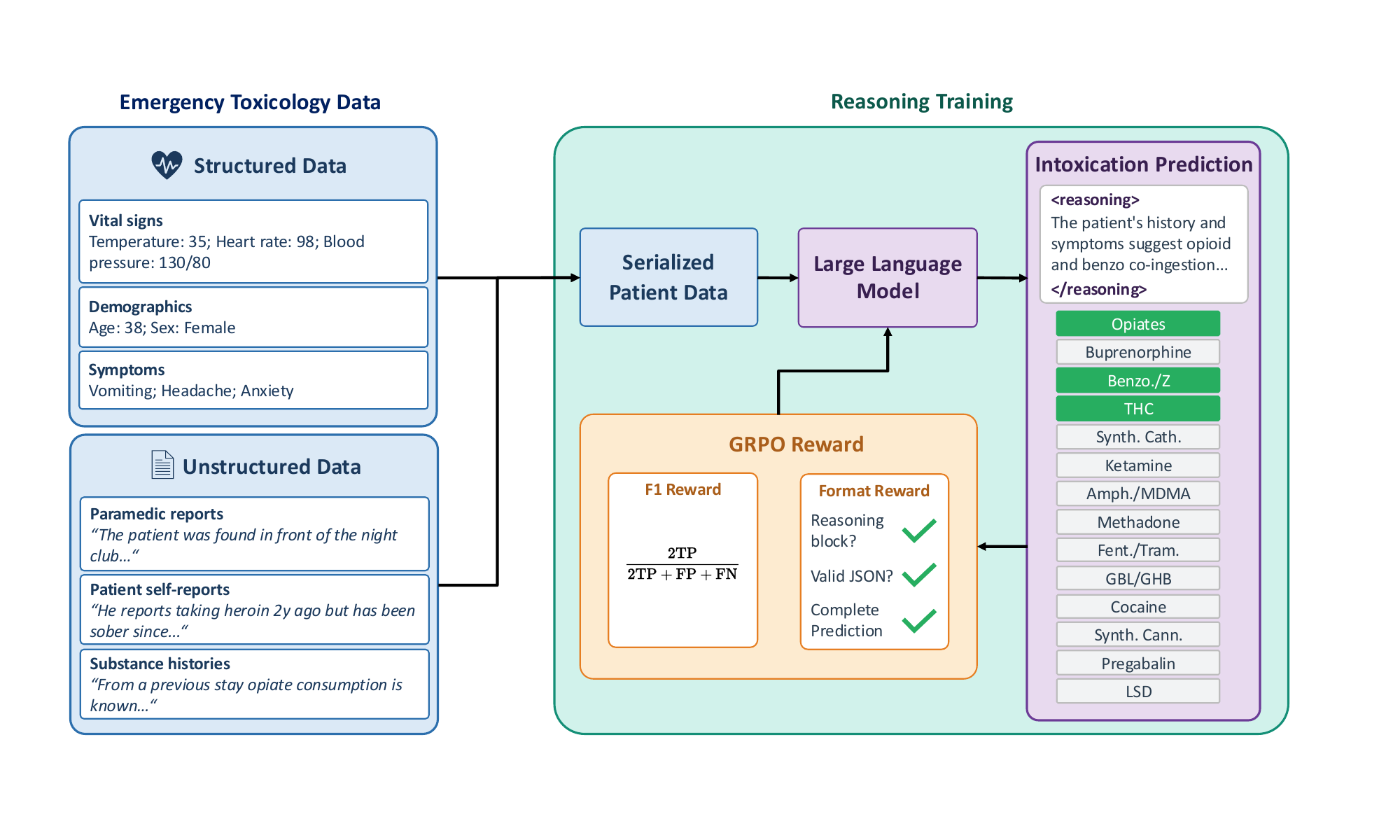}
    \caption{Overview of DeToxR. Heterogeneous emergency-department data is combined and used to prompt an LLM for diagnostic reasoning and multi-label toxin prediction. The model is trained with GRPO using an F1 and format reward.}
    \label{fig:method_overview}
\end{figure}

Early computational approaches have relied on probabilistic logic networks~\cite{chary2021diagnosis} and gradient boosting~\cite{mehrpour2022classification}, while more recent methods have employed Deep Neural Networks~\cite{mehrpour2023deep} or Graph Neural Networks~\cite{burwinkel2020decision,zellner2023toxnet} being limited to structured numerical inputs. Large Language Models (LLMs) are well-suited to the clinical toxicology setting because they can process both unstructured narrative reporting and structured clinical variables in a single forward pass, closing the modality gap that has constrained prior approaches to structured inputs alone. To shape LLM behavior for complex reasoning tasks, reinforcement learning-based post-training, specifically Group Relative Policy Optimization (GRPO), has emerged as a promising approach. By optimizing directly for verifiable objectives, models can be trained to reason before providing their final prediction without the need for ground-truth reasoning traces~\cite{guo2025deepseek}. Recent work has begun exploring GRPO for medical diagnosis~\cite{zamai2025explainable,lai2026med,pan2502medvlm,bani2025language}. However, these existing applications operate in solely clinical environments, relying on medical data without incorporating any non-clinical context and utilize binary accuracy rewards to judge performance.\looseness=-1

In this work, we present DeToxR (Decision-support for Toxicology with Reasoning), the first adaptation of GRPO to the highly heterogeneous domain of emergency toxicology. We use a data-fusion formulation to target multi-label prediction of poly-intoxications across 14 substance classes, jointly reasoning over structured clinical variables and unstructured narratives, including patient histories and paramedic reports. We align lightweight LLMs with a task-specific clinical performance reward that explicitly penalizes both missing co-ingestants and hallucinated poisons, while encouraging structured diagnostic reasoning. Finally, we benchmark our model against various baselines and a medical expert in a preliminary clinical validation study, showing promising diagnostic performance.\looseness=-1

%% file: sections/02_methodology.tex
\subsubsection{Problem formulation.}
We formalize emergency poly-substance intoxication prediction as a multi-label classification problem. Each case $i$ is represented by a tuple $x_i = (s_i, t_i)$ of structured clinical variables $s_i$ and unstructured free-text $t_i$, with the prediction target being a binary vector $y_i \in \{0,1\}^K$ over $K$ substance classes, determined by retrospective laboratory confirmation.

\subsubsection{Heterogeneous data fusion.}
DeToxR serializes all available evidence into a single Markdown-formatted natural-language prompt. Structured clinical variables, including demographics (age, sex), vital signs (e.g. heart rate, blood pressure), binary clinical indicators (e.g. coma on admission, preclinical intubation) and symptom flags (e.g. vomiting, seizures), are converted to readable key-value pairs. Numeric values are directly included, with missing entries represented as "N/A", while binary indicators are mapped to boolean values. Unstructured free-text fields, including the patient history, physical examination report, and ECG findings, are inserted into dedicated sections of the prompt without any processing. When available, the substances contained in the history are included as a bulleted list.\looseness=-1

\subsubsection{Output schema.}
We instruct the model to output a JSON object with binary predictions for all $K$ substances, preceded by a diagnostic reasoning trace enclosed in \texttt{<reasoning>} tags. This provides both machine-verifiable predictions and clinician-readable rationales.\looseness=-1

\subsubsection{Reinforcement learning finetuning.}
We finetune the LLM using GRPO~\cite{guo2025deepseek} with DAPO loss aggregation~\cite{yu2025dapo} and sequence-level importance sampling~\cite{zheng2025group}. We propose a composite reward $R = R_{\text{F1}} + R_{\text{format}}$, where the task reward is the sample-level F1 score, calculated from true positives (TP), false positives (FP), and false negatives (FN):

\begin{equation}
R_{\text{F1}} = \begin{cases} \frac{2\text{TP}}{2\text{TP} + \text{FP} + \text{FN}}, & \text{TP} + \text{FP} + \text{FN} > 0 \\ 1, & \text{otherwise} \end{cases}
\end{equation}

F1 penalizes missed substances and hallucinated poisons symmetrically relative to correct identifications. This is particularly relevant in the poly-intoxication setting, where the number of active substances varies across cases and the label vectors are typically sparse.\looseness=-1

The format component $R_{\text{format}}$ rewards valid structured outputs: a reasoning block (+0.25), a valid JSON block (+0.25), and a correctly keyed JSON object containing all $K$ substance entries (+0.5), yielding a maximum format reward of 1.0. Together with $R_{\text{F1}}$, this drives the model to produce accurate multi-label predictions, parseable outputs, and diagnostic reasoning traces without requiring ground-truth rationales.\looseness=-1

%% file: sections/03_experiments.tex
\begin{table}[t]
\centering
\caption{Evaluation results for simple baselines, zero-shot inference on various base LLMs, and finetuned methods including DeToxR. Higher is better for all metrics. Best results are marked in bold. Second best are underlined. 95\% confidence intervals are provided in square brackets.}
\label{tab:main_results}
\begin{adjustbox}{width=\textwidth}
\small
\setlength{\tabcolsep}{3pt}
\renewcommand{\arraystretch}{1.2}
\begin{tabular}{l l c c c c c c}
\toprule
\multirow{2}{*}{\textbf{Category}} & \multirow{2}{*}{\textbf{Method}} & \multicolumn{2}{c}{\textbf{F1 (\%)}} & \multicolumn{2}{c}{\textbf{Recall (\%)}} & \multicolumn{2}{c}{\textbf{Precision (\%)}} \\
\cmidrule(lr){3-4} \cmidrule(lr){5-6} \cmidrule(lr){7-8}
& & \textbf{Micro} & \textbf{Macro} & \textbf{Micro} & \textbf{Macro} & \textbf{Micro} & \textbf{Macro} \\
\midrule
\multirow{3}{*}{Baselines} & History baseline & 48.8 [44.3, 53.2] & 53.5 [43.2, 57.8] & 36.1 [32.0, 40.5] & 48.7 [38.7, 53.2] & \textbf{75.3} [70.1, 80.2] & \textbf{73.4} [63.3, 79.3] \\
                            & MLP & 57.6 [53.9, 61.1] & 36.0 [31.6, 39.6] & 47.5 [43.6, 51.5] & 29.5 [26.1, 33.2] & \underline{72.9} [68.6, 77.3] & 52.7 [45.6, 57.3]  \\
                            & XGBoost & 56.4 [52.7, 59.9] & 34.6 [30.6, 38.2] & 49.4 [45.2, 53.6] & 30.2 [27.0, 33.7] & 65.7 [61.6, 69.7] & 44.4 [38.3, 51.3] \\
\midrule
\multirow{6}{*}{\makecell{Base LLM\\inference}} &
Qwen3 4B Instruct & 52.6 [49.1, 56.0] & 50.8 [44.2, 55.4] & 45.8 [42.2, 49.6] & 56.0 [46.4, 59.9] & 61.8 [57.6, 65.8] & 57.8 [52.1, 63.5] \\
& Qwen3 4B Thinking & 48.2 [44.4, 52.1] & 46.2 [39.5, 50.9] & 38.2 [34.5, 42.1] & 49.4 [39.7, 53.5] & 65.3 [60.7, 69.8] & 59.8 [53.8, 65.8] \\
& Llama 3.1 8B Instruct & 49.8 [46.2, 53.4] & 48.9 [42.0, 53.8] & 50.1 [46.4, 53.9] & 59.6 [49.8, 63.7] & 49.5 [45.1, 54.0] & 50.0 [44.1, 56.2] \\
& Qwen3 30B (MoE) & 50.1 [46.8, 53.6] & 47.3 [41.8, 51.8] & 50.2 [46.5, 54.1] & \textbf{60.7} [51.0, 64.8] & 50.1 [46.0, 54.2] & 49.6 [44.8, 54.5] \\
& GPT OSS 20B & 48.8 [45.2, 52.4] & 47.6 [40.2, 52.9] & 38.9 [35.3, 42.7] & 48.4 [38.0, 53.9] & 65.3 [60.5, 70.0] & 59.9 [52.6, 68.0] \\
& MedGemma 4B IT & 50.6 [46.9, 54.2] & 52.4 [43.7, 56.8] & 43.9 [39.9, 48.0] & 54.6 [44.7, 58.9] & 59.8 [55.5, 64.2] & 60.0 [52.3, 66.5] \\
\midrule
\multirow{2}{*}{\makecell{Finetuned\\models}} &
Qwen3 4B Inst. + SFT & \underline{58.9} [55.7, 62.1] & \underline{55.3} [45.5, 59.3] & \underline{52.3} [48.8, 56.0] & 52.1 [42.6, 55.8] & 67.3 [63.1, 71.6] & \underline{68.4} [57.1, 75.1] \\
&
DeToxR & \textbf{63.7} [60.8, 66.7] & \textbf{59.9} [50.0, 64.0] & \textbf{63.9} [60.8, 67.2] & \underline{60.2} [50.6, 64.1] & 63.5 [59.6, 67.5] & 66.9 [57.0, 72.7] \\
\bottomrule
\end{tabular}
\end{adjustbox}
\end{table}

\subsubsection{Experimental setup.} The dataset stems from the toxicology department of a single hospital in Munich, Germany and contains a total of 870 poly-in\-tox\-i\-ca\-tion cases, collected between 2014 and 2022. The cohort includes adult patients with a median age of 33 years and a sex distribution of 74.14\% male and 25.86\% female. The patients were screened for the following $K=14$ substances: Opiates (\(n=251\)), methadone (\(n=193\)), bu\-pre\-nor\-phine (\(n=157\)), fentanyl\slash tramadol (\(n=67\)), ben\-zo\-di\-az\-e\-pines\slash z-sub\-stances (\(n=465\)), GBL\slash GHB (\(n=32\)), THC (\(n=329\)), cocaine (\(n=146\)), NPS\slash cathi\-nones (\(n=70\)), synthetic cannabinoids (\(n=53\)), ketamine (\(n=4\)), pregabalin (\(n=384\)), am\-phet\-a\-mines\slash MDMA (\(n=187\)) and LSD (\(n=9\)). The average number of present toxins per case is 2.7. The dataset contains a variety of structured clinical data: patient age and sex, vital signs, symptoms and the substance history. In our setting, the substance history describes the substances which were reportedly consumed by the patient. Additionally, there are three fields with text data: one containing the results of a physical examination, one with ECG findings, and one with the detailed background history of the patient's admission. In order to retain a sufficiently large test set, we split the data into train, validation and test partitions using a 50:20:30 ratio. To obtain balanced splits in the multi-label setting, we use iterative stratification~\cite{sechidis2011stratification}, which aims to preserve the distribution of label evidence and label co-occurrence relations across splits.\looseness=-1

\begin{table}[t]
\centering
\caption{Results of the ablation studies. We evaluate a different reward formulation, and employing our method on another base model.}
\label{tab:ablation_results}
\begin{adjustbox}{width=\textwidth}
\small
\setlength{\tabcolsep}{3pt}
\renewcommand{\arraystretch}{1.2}
\begin{tabular}{l l c c c c c c}
\toprule
\multirow{2}{*}{\textbf{Category}} & \multirow{2}{*}{\textbf{Method}} & \multicolumn{2}{c}{\textbf{F1 (\%)}} & \multicolumn{2}{c}{\textbf{Recall (\%)}} & \multicolumn{2}{c}{\textbf{Precision (\%)}} \\
\cmidrule(lr){3-4} \cmidrule(lr){5-6} \cmidrule(lr){7-8}
& & \textbf{Micro} & \textbf{Macro} & \textbf{Micro} & \textbf{Macro} & \textbf{Micro} & \textbf{Macro} \\
\midrule
\multirow{2}{*}{Reward formulation} 
& DeToxR (IoU reward) & 61.1 [57.7, 64.5] & 55.7 [47.5, 60.6] & 57.5 [53.9, 61.2] & 57.8 [48.2, 61.7] & \textbf{65.2} [60.7, 69.5] & 61.1 [53.8, 68.3] \\
& DeToxR (F1 reward) & \textbf{63.7} [60.8, 66.7] & \textbf{59.9} [50.0, 64.0] & \textbf{63.9} [60.8, 67.2] & \textbf{60.2} [50.6, 64.1] & 63.5 [59.6, 67.5] & \textbf{66.9} [57.0, 72.7] \\
\midrule
\multirow{2}{*}{Other LLM backbone} &
Llama 3.2 3B Instruct & 52.7 [49.2, 56.3] & 49.2 [43.1, 53.5] & 50.4 [46.5, 54.5] & \textbf{60.1} [50.3, 64.2] & 55.3 [51.1, 59.7] & 52.0 [46.9, 57.2] \\
& Llama 3.2 3B Instruct (DeToxR) & \textbf{57.9} [54.7, 61.1] & \textbf{57.0} [47.0, 60.9] & \textbf{51.5} [48.0, 55.1] & 56.1 [46.2, 60.1] & \textbf{66.1} [61.9, 70.3] & \textbf{68.3} [58.5, 73.6] \\
\bottomrule
\end{tabular}
\end{adjustbox}
\end{table}

\subsubsection{Baselines.} As a simple reference point, we use a \emph{history baseline}, where the provided substance history is treated as the prediction. Additionally, we include two classical machine learning baselines: an MLP and an XGBoost-based multi-output classifier. Both methods are trained only on structured data as unstructured data cannot be natively processed by these methods. The MLP uses three hidden layers with sizes \(640\), \(320\), and \(64\). The XGBoost classifier is trained per label via a multi-output wrapper, fitting on the same scaled features used for the MLP. We also evaluate a range of open-weight LLMs with different numbers of parameters, reasoning capabilities and areas of expertise, and report results for the following set of models: Qwen3 4B Instruct 2507, Qwen3 4B Thinking 2507, Llama 3.1 8B Instruct, Qwen3 30B A3B Instruct 2507, GPT OSS 20B, and MedGemma 4B IT.\looseness=-1

\subsubsection{Finetuned models.} We conduct supervised finetuning (SFT) to predict the target labels directly. For that, we train using a linearly scheduled learning rate of \(2 \cdot 10^{-4}\), AdamW optimizer, weight decay of \(0.001\), and LoRA~\cite{hu2022lora}. For DeToxR, trained with GRPO, we again employ LoRA and train with learning rate \(2 \cdot 10^{-5}\) and a group size of 3. Training runs for at most 10,000 steps, we evaluate and save checkpoints every 500 steps, and select the final model as the checkpoint with the highest validation score. \looseness=-1

All finetuning experiments are conducted with Qwen3 4B Instruct as the base model. This model was chosen due to its favorable balance of a small parameter count and strong capabilities. All models were trained on a single NVIDIA A40 GPU. At inference time, DeToxR required approximately 9 seconds per sample on this device.\looseness=-1

%% file: sections/04_results.tex
\subsubsection{Performance comparison against other methods.} Table~\ref{tab:main_results} summarizes the performance of all baselines as well as our SFT and GRPO finetuned models across micro- and macro-averaged recall and precision, and F1 score, which is their harmonic mean reflecting the overall performance. Compared to the history baseline, DeToxR shows a strong improvement in F1 score and recall, while the history baseline achieves the highest precision. This is expected as the history baseline is based on patient-reported poison exposures. Patients seldom report drugs they did not actually consume, but they may fail to mention substances due to fear of legal consequences or altered mental states. 

Relative to the other non-LLM baselines, we observe a strong gain in F1, recall, and macro-precision. Against the base LLMs, DeToxR achieves higher F1, higher micro-recall and higher macro-precision, while matching macro-recall. In micro-precision, DeToxR is slightly outperformed by the two reasoning models Qwen3 4B Thinking and GPT OSS 20B. 

Finally, when comparing DeToxR with an SFT finetuned baseline, DeToxR achieves a large improvement in F1 and recall with a smaller decrease in precision. Due to the low prevalence of positive labels, the SFT method struggles to predict positive poisonings, causing higher precision at the cost of much lower recall.

Paired bootstrap comparisons on the shared test cases (10,000 resamples) show that DeToxR significantly outperforms every Table~\ref{tab:main_results} comparator in micro- and macro-F1. Differences were deemed significant when the two-sided 95\% confidence interval for $\Delta$ excluded zero.

\begin{figure}[t]
    \centering
    \includegraphics[width=\linewidth]{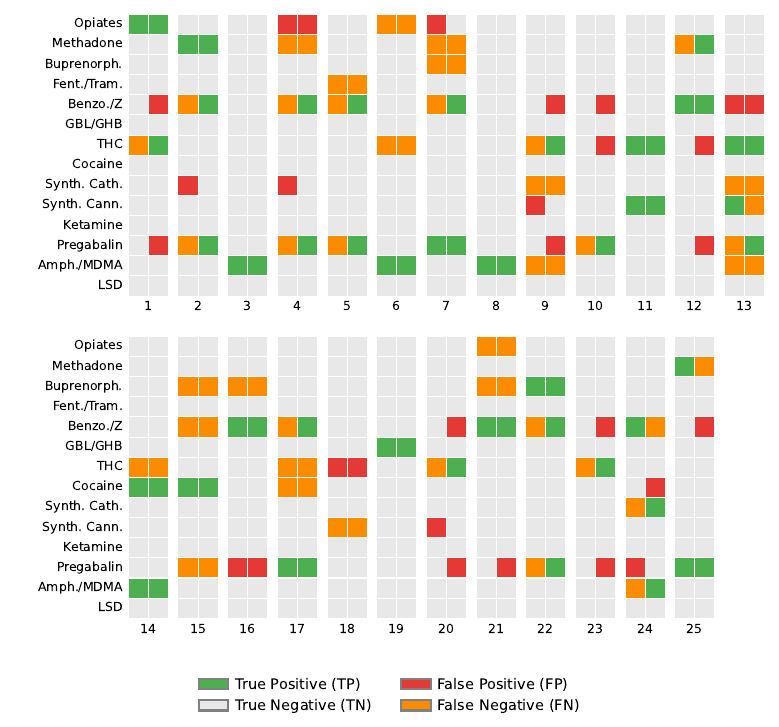}
    \caption{Comparison of the toxicologist against our model on 25 randomly selected test set cases. For each case, the left column is the toxicologist's prediction, and the right one is the prediction of DeToxR.}
    \label{fig:case_level_comparison}
\end{figure}

\subsubsection{Performance comparison against a medical expert.} In order to evaluate the performance of DeToxR against a medical expert, we conducted a small survey with one toxicologist from a hospital's toxicology department, who was tasked to make predictions for 25 randomly selected intoxication cases from the test set. In this initial comparison, our method achieves a higher F1 score than the toxicologist, with micro-F1 of 0.644 vs 0.473 and macro-F1 of 0.488 vs 0.398. The toxicologist perfectly predicts 5 of 25 cases, while DeToxR perfectly predicts 6 cases. There are 19 intoxications that are correctly predicted by DeToxR while being missed by the toxicologist, and only 3 cases where the toxicologist identifies a toxin not found by DeToxR. However, our method makes false positive predictions in 15 cases where the toxicologist correctly predicts no intoxication, while there are only 6 cases where it is the other way around. This behavior is supported by the fact that the toxicologist's precision of 0.688 micro and 0.564 macro is better to that of our model, with 0.667 micro and 0.640 macro, while the toxicologist's recall of 0.361 micro and 0.337 macro is significantly lower than that of DeToxR, with 0.623 micro and 0.455 macro. A paired Wilcoxon signed-rank test across the 25 cases showed that DeToxR's improvements are statistically significant for recall and micro-F1, whereas the difference in micro-precision is small and not significant. Figure \ref{fig:case_level_comparison} displays a detailed case-level comparison between the toxicologist and DeToxR.\looseness=-1

\input{figures/reasoning_trace}

\subsubsection{Ablations.} We evaluate the impact of the reward function by comparing our proposed F1-based reward against an Intersection over Union (IoU) alternative, which for the set $P = \{k \vert \hat{y}_k = 1\}$ of positive predictions and the set $G = \{k \vert y_k = 1\}$ of positive ground truths is defined as
\begin{equation}
R_{\text{IoU}} = \begin{cases}
    1.0, & \quad \text{if}\ P=G=\emptyset\\
    \frac{\lvert P \cap G\rvert}{\lvert P \cup G\rvert}, & \quad \text{otherwise}
\end{cases}
\end{equation}
The F1-reward model outperforms the IoU-reward variant, highlighting the importance of good, task-specific reward design. 

To demonstrate the broader applicability of our GRPO finetuning approach, we apply it to Meta's Llama 3.2 3B Instruct, where our finetuning strategy yields strong performance gains over the base model.


\subsubsection{Qualitative reasoning analysis.}
In Figure \ref{fig:reasoning_trace} we show an exemplary reasoning trace of DeToxR. It shows that the model correctly analyses clinical findings and successfully reasons based on a combination of structured and unstructured findings. However, we also observe that the model sometimes tends to rationalize its correct predictions by hallucinating non-existent findings that would support these prediction instead of basing its reasoning on the actual patient information. Therefore, reasoning should be primarily regarded as an internal step that aims to improve predictive performance and caution should be exercised when providing reasoning traces to clinicians for decision support. If done correctly, exposition to the potential inconsistencies could however help them get a better understanding of the model's limitations.\looseness=-1

%% file: figures/reasoning_trace.tex
\begin{figure}[t]
    \centering
    \begin{reasoningbox}
        {[...]} The key elements guiding the diagnosis are:
        \begin{itemize}
            \item \textcolor{green!60!black}{Known heroin abuse}
            \item \textcolor{green!60!black}{Absence of other substances in the toxicological evaluation}
            \item \textcolor{green!60!black}{Bradypnoea and hypoxaemia requiring high-flow oxygen support}
            \item \textcolor{red!60!black}{ECG findings consistent with toxicological cardiopulmonary effects rather than specific drug-related abnormalities}
            \item \textcolor{green!60!black}{Normal cardiac rhythm and absence of ERBS or conduction disturbances}
        \end{itemize}
        \textcolor{green!60!black}{The ECG findings [...] are non-specific} \textcolor{red!60!black}{and may be attributable to the toxicological burden or hypoxia. The absence of block or ERBS supports a toxicological rather than cardiac etiology}.\\\\
        \textcolor{green!60!black}{The presence of miotic pupils, reduced AZ, and preserved PSR is highly suggestive of opiate intoxication. The absence of other substances in the list, coupled with the known heroin use, strongly supports an opiate intoxication. The critical question is whether specific opiates [...] or additional substances (e.g., Z-substances) contribute to the respiratory} and \textcolor{red!60!black}{cardiovascular instability}.\\\\
        \textcolor{red!60!black}{The high lactate (1.4)} and \textcolor{green!60!black}{marked respiratory compromise suggest a toxicological respiratory depression, which is consistent with opiates.} \textcolor{red!60!black}{The absence of other substances such as benzodiazepines or stimulants reduces the likelihood of contributing factors.} \textcolor{green!60!black}{However, the presence of Z-substances must be considered, especially given the respiratory} and \textcolor{red!60!black}{cardiovascular instability}. [...]\\\\
        \textcolor{green!60!black}{Pregabalin intoxication is a well-documented contributor to respiratory depression} and \textcolor{red!60!black}{cardiovascular instability} \textcolor{green!60!black}{in the setting of opiate intoxication}. 
        [...]
    \end{reasoningbox}
    \caption{Excerpts from an example reasoning trace. The patient of the underlying case is intoxicated with opiates, benzodiazepines/z-substances and pregabalin, which DeToxR correctly predicts. Claims in green are supported by the provided data or logically sound, while red ones are not supported or wrong.}
    \label{fig:reasoning_trace}
\end{figure}

%% file: sections/05_conclusion.tex
We present DeToxR, the first adaptation of reinforcement learning to the challenging domain of emergency toxicology, using GRPO to finetune lightweight LLMs for multi-label substance prediction across 14 toxin classes. By optimizing directly for a clinical F1 reward, our approach learns to fuse unstructured free-text narratives with structured clinical data and improves over classical baselines, zero-shot LLMs, and supervised finetuning. In a preliminary comparison with one expert toxicologist, DeToxR achieved higher F1 and recall. \\
\textbf{Prospect of application.} The proposed method could potentially be used to provide decision support to clinicians in toxicology emergency care by processing narrative reports and clinical parameters to propose early diagnoses for confirmation by the toxicologist.